\documentclass[10pt,twocolumn,letterpaper]{article}

\usepackage{cvpr}              

\usepackage{xcolor}
\usepackage{amssymb}
\usepackage{booktabs}  
\usepackage{multirow}  
\usepackage[normalem]{ulem} 
\usepackage{float}
\usepackage{graphicx} 
\usepackage{bm} 
\usepackage{makecell}

\definecolor{cvprblue}{rgb}{0.21,0.49,0.74}
\usepackage[pagebackref,breaklinks,colorlinks,allcolors=cvprblue]{hyperref}

\def\paperID{38630} 
\def\confName{CVPR}
\def\confYear{2026}

\title{Select, Hypothesize and Verify: Towards Verified Neuron Concept Interpretation}

\author{
  ZeBin Ji\footnotemark[2]~~\textsuperscript{\rm, 1}, ~
  Yang Hu\footnotemark[2]~~\textsuperscript{\rm, 1}, ~
  Xiuli Bi\footnotemark[1]~~\textsuperscript{\rm, 1}, ~
  Bo Liu \textsuperscript{\rm 1}, ~
  Bin Xiao\textsuperscript{\rm 1, 2} \\
  \textsuperscript{\rm 1}Chongqing Key Laboratory of Image Cognition,\\ Chongqing University of Posts and Telecommunications, Chongqing, China\\
  \textsuperscript{\rm 2}Jinan Inspur Data Technology Co., Ltd., Jinan, China\\
  {\tt\small \{S230201044, S210201038\}@stu.cqupt.edu.cn}, {\tt\small \{bixl, boliu, xiaobin\}@cqupt.edu.cn}
}

\begin{document}
\twocolumn[{
\maketitle
\vspace{-16pt} 

\begin{figure}[H]  
\hsize=\textwidth 
  \centering
  \includegraphics[width=0.98\textwidth]{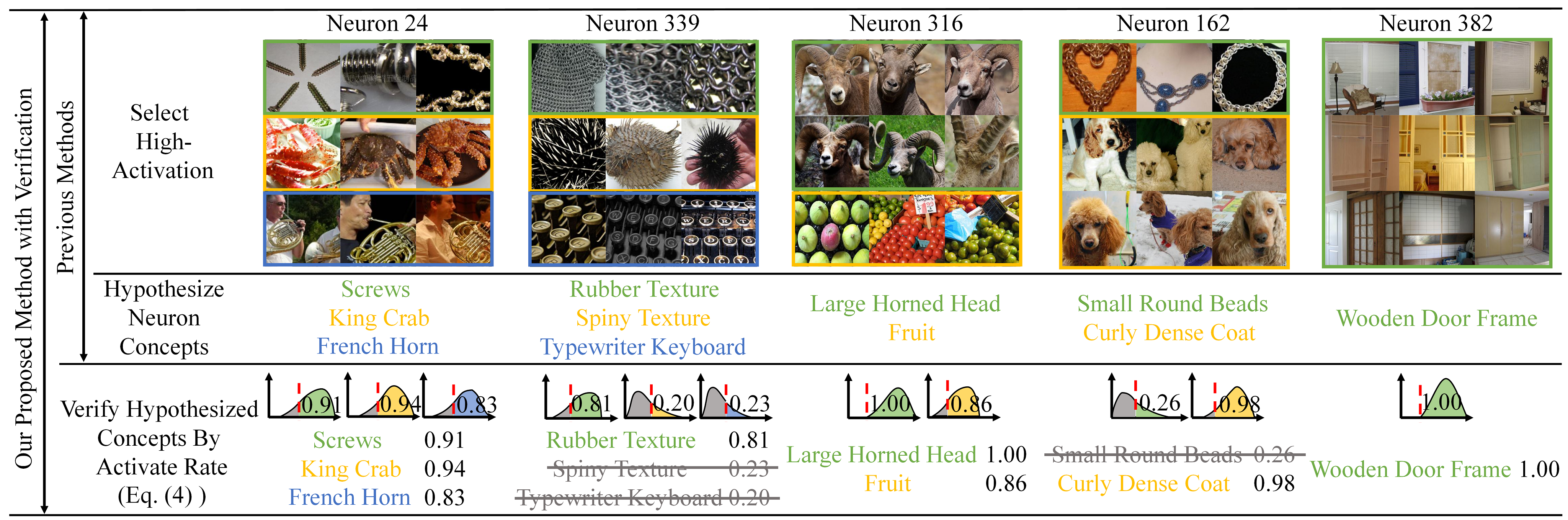}  
  \caption{\textbf{Comparison of previous methods and our proposed method with verification.} The neurons are sampled from the second-to-last layer of ResNet-18. Existing methods assume that concepts inferred from activated neurons are all accurate, while our proposed method introduces verification to identify incorrect ones. Specifically, for neuron 162, both the concepts (Small Round Beards and Curly Dense Coat) are inferred from high-activation images. Verification shows that the activation rate for small round whiskers (0.26) suggests this concept may be incorrect, while curly dense hair achieves 0.98, confirming the correctness of the concept.
  }
  \label{fig:main1}
\end{figure}
}]
\footnotetext[2]{These authors contributed equally}  
\footnotetext[1]{Corresponding Author}               


\begin{abstract}
It is essential for understanding neural network decisions to interpret the functionality (also known as concepts) of neurons. Existing approaches describe neuron concepts by generating natural language descriptions, thereby advancing the understanding of the neural network's decision-making mechanism. However, these approaches assume that each neuron has well-defined functions and provides discriminative features for neural network decision-making. In fact, some neurons may be redundant or may offer misleading concepts. Thus, the descriptions for such neurons may cause misinterpretations of the factors driving the neural network’s decisions. To address the issue, we introduce a verification of neuron functions, which checks whether the generated concept highly activates the corresponding neuron. Furthermore, we propose a Select–Hypothesize–Verify framework for interpreting neuron functionality. This framework consists of: 1) selecting activation samples that best capture a neuron’s well-defined functional behavior through activation-distribution analysis; 2) forming hypotheses about concepts for the selected neurons; and 3) verifying whether the generated concepts accurately reflect the functionality of the neuron. Extensive experiments show that our method produces more accurate neuron concepts. Our generated concepts activate the corresponding neurons with a probability approximately 1.5 times that of the current state-of-the-art method.

\end{abstract}


\section{Introduction}
\label{sec:intro}
Deep neural networks (DNNs) have achieved remarkable progress in various domains. However, due to their complex structures, understanding the decision-making process of DNNs remains challenging. This opacity not only undermines trust in the models but also limits their deployment in safety-critical applications. To address this issue, explainable artificial intelligence (XAI) has emerged, aiming to reveal the internal mechanisms of neural networks and make their decision processes transparent and comprehensible.


\newcommand{\good}{\textcolor{blue}{\textbf{$\checkmark$}}}
\newcommand{\bad}{\textcolor{red}{\textbf{$\times$}}}
\newcommand{\partialy}{\textcolor{orange}{\textbf{$\triangle$}}}


Previous studies have attempted to break down convolutional neural networks (CNNs) into interpretable neurons, using the visualization techniques~\cite{simonyan2013deep,zeiler2014visualizing,mahendran2015understanding,zhou2016learning,olah2017feature} (e.g., Saliency Maps~\cite{simonyan2013deep} and Grad-CAM~\cite{selvaraju2017grad}) to understand models from the most fundamental neurons (also called units, filters, or features). Visualization techniques help to reveal the functions reflected by neurons, but noisy activation can make them challenging to interpret without manual inspection. To address this limitation, researchers have begun describing the functionality of neurons (also referred to as concepts) using natural language, such as Network Dissection~\cite{bau2017network} and Compositional Explanations~\cite{mu2020compositional}. Network Dissection constructs the Broden probe dataset to establish correspondences between visualizations and predefined concepts, while Compositional Explanations extends this approach to generate complex concept compositions. CLIP-Dissect~\cite{Dissect} further leverages OpenAI’s CLIP~\cite{radford2021learning} model to achieve more flexible matching. 

These works analyze the entire network, providing functional descriptions of neurons in layers close to the input that primarily focus on visual features, such as color and texture. Although these features are indispensable for the model’s decision-making, they contribute only to a limited extent to human understanding of the neural network. Consequently, subsequent research has shifted attention toward neurons in the penultimate layer, whose functions are more aligned with human-understandable concepts. For example, FALCON~\cite{kalibhat2023identifying} focuses exclusively on the final layer, removing spurious descriptions by contrasting low-level activations with the corresponding images. Following FALCON, WWW~\cite{ahn2024unified} improves the accuracy of neuron descriptions by integrating activation locations and contributions to predictions. DnD~\cite{bai2025interpreting} uses large language models to generate natural language descriptions, providing higher-quality concepts for neurons.




Although these methods aim to describe neuron functionality accurately, they all rely on the same assumption: that each neuron has a well-defined function and provides discriminative features for neural network decision-making. However, existing studies~\cite{nanda2023diffused,dalvi2020analyzing} have shown that networks contain redundant neurons that do not contribute to decisions. Describing such neurons may be misleading by interpreting noisy activations as meaningful functionality, thereby misleading human understanding of neural network decision mechanisms. Moreover, current approaches typically infer neuronal functions based on their activation distributions across probe datasets, essentially involving an observational hypothesis process. However, due to the limited coverage of probe data, these hypotheses may be subject to dataset biases and may not accurately reflect the roles of the neurons. Therefore, we argue that intervention-based experiments are necessary to validate these hypotheses.

This principle aligns with scientific methodology: scientific understanding arises not only from observation but also from formulating hypotheses and validating them through controlled experiments~\cite{poincare1914science,popper1959logic}. For a long time, neuroscience research has followed an “Observe–Hypothesize–Verify” paradigm: researchers first record neural activity to identify neurons associated with specific functions (observation and localization); then propose functional hypotheses based on response patterns (hypothesis); and finally verify the causal role of these neurons in behavior or cognition through controlled experiments (verification). This closed-loop logic has driven progress in fields such as the visual and language cortices, providing a reliable methodology for understanding the brain.

Similarly, we would argue that research into the interpretability of deep neural networks should adhere to the same scientific logic. Inspired by this idea, we propose a Select–Hypothesize–Verify (SIEVE) framework for interpreting neuron functionality. The framework consists of: 1) Select: Identify data samples that consistently elicit high activation for a given neuron based on its activation distribution over the probe dataset; 2) Hypothesize: formulate hypotheses about the neuron’s functional role using these consistently high-activation samples; 3) Verify: conduct intervention experiments on the neuron to assess the strength of the association between the hypothesized concepts and the neuron’s function, thereby ensuring that the hypotheses reliably reflect the internal mechanisms of the neural network.
Our main contributions are as follows:
\begin{itemize}
    \item Existing methods are limited by the assumption that concepts extracted from visual data by large models must be accurate. To mitigate this issue, we propose a Select–Hypothesize–Verify framework for interpreting neuron functionality (dubbed SIEVE), achieving closed-loop verification to remove mismatched concepts.
    \item This paper demonstrates that some neurons may not provide discriminative features for neural network decision-making. We design a neural filtering mechanism to prevent the introduction of erroneous concepts.
    \item Extensive experiments validate the effectiveness of our framework, and the generated concepts activate the corresponding neurons with a probability close to 1.5× that of the current state-of-the-art method.
\end{itemize}





\section{Related Works}
\label{sec:Related}


\textbf{Visualization.} Interpretability is defined as the ability to provide explanations that are understandable to humans.~\cite{doshi2017towards,zhang2021survey}. 
Early studies mainly achieved this goal through visualization techniques. Representative methods include Saliency Maps~\cite{simonyan2013deep} and Grad-CAM~\cite{selvaraju2017grad}, which highlight the regions most important for predictions by computing the gradient of the network output with respect to the input pixels. This helps humans to understand which parts of the input the model focuses on. Additionally, activation maximization~\cite{mordvintsev2015deepdream,olah2017feature,nguyen2016synthesizing} optimizes the input image to maximize the activation of a neuron, thereby revealing its functionality. Although visualization aids human understanding of neural networks, manual inspection is typically required to determine whether the visualization results are reasonable.

\textbf{Natural Language Explanations.} To provide more explicit explanations, researchers have begun to describe neuron functionality in natural language.
Network Dissection~\cite{bau2017network} pioneered this direction by constructing Broden, a pixel-wise annotated dataset, and establishing correspondences between neurons and concepts through the intersection over union score (IoU) of neuron activation maps and concept segmentation maps. 
Subsequently, Bau et al.~\cite{bau2020understanding} utilized segmentation models to automatically annotate datasets; however, the concept set remained limited by the capacity of pre-trained segmentation models.
MILAN~\cite{hernandez2021natural} trained image-to-text models to generate unrestricted descriptions.
CLIP-Dissect~\cite{Dissect} and FALCON~\cite{kalibhat2023identifying} further utilized CLIP to describe neuron functionality without annotations. In contrast, DnD~\cite{bai2025interpreting} employed multimodal models to produce more complex natural language descriptions. Bills et al.~\cite{bills2023language} proposed using LLMs to predict neuron activations of input. MAIA~\cite{shaham2024multimodal} introduced a multimodal agent that autonomously composes API-based tools to conduct iterative experiments for interpreting neurons in vision models. These methods focus on highly activated neuron samples, ignoring the possibility that such activations may merely correspond to average levels and that the neurons may not have well-defined functions. Although Rosa et al.~\cite{la2023towards} and Oikarinen \& Weng.~\cite{oikarinen2024linear} attempted to provide a more comprehensive description of each neuron by subdividing activation ranges, low activation may lead to ambiguous interpretations. By contrast, we focus not only on highly activated samples but also consider whether the neurons provide discriminative features for decision-making.



\textbf{Intervention-Based Verification.} Intervention-based verification methods typically investigate the causal relationship between internal representations and final model decisions by actively modifying input samples or the internal state of the model, such as the output of a single neuron. Early works, such as LIME~\cite{ribeiro2016should} and Zeiler \& Fergus.~\cite{zeiler2014visualizing}, applied local perturbations in the input space to assess the impact of different image regions on model predictions. Morcos et al.~\cite{morcos2018importance} further conducted interventions within the network by setting neuron activations to zero, analyzing the importance of individual neurons for model performance. However, these intervention-based verification approaches primarily aim to identify which inputs or neurons the model relies on, and do not evaluate whether natural language descriptions accurately correspond to the functionality of these neurons. Consequently, they cannot answer whether a neuron truly encodes the intended concept. To address this limitation, we propose an intervention verification method driven by hypothesized concepts. 
This approach involves constructing input samples based on hypothesized concepts, and then evaluating the consistency between these concepts and the true functionality of neurons by carrying out targeted interventions and observing changes in neuron activation.

\begin{figure*}[t]  
  \centering
  \includegraphics[width=0.95\textwidth]{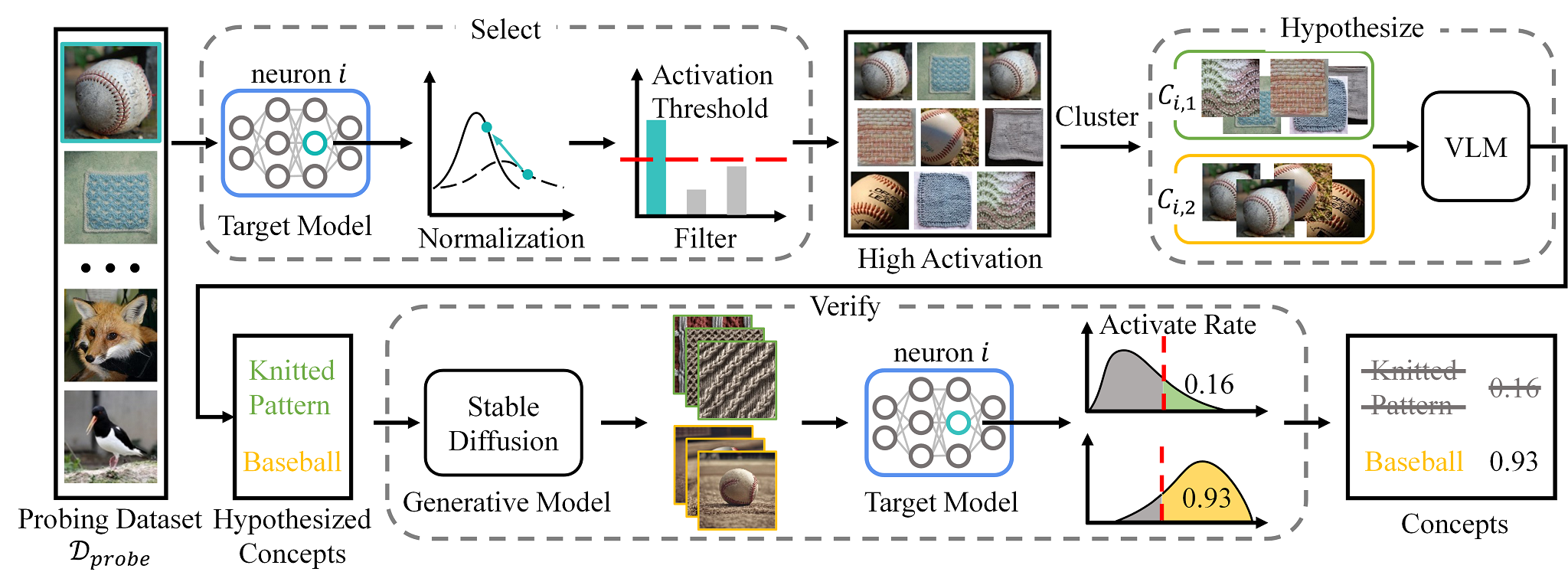}  
  \caption{
  \textbf{Overview of SIEVE.} 1) Select: For each sample, we obtain the normalized activation of neuron $i$ from the network’s penultimate layer. Samples with activations above a predefined threshold are identified as high-activation samples. 2) These high-activation samples are clustered, and a vision-language model is used to generate the hypothesized concepts for each cluster. 3) Each concept is converted into a semantic image using text-to-image generation. The generated images are fed into the target model to measure the activation rate of neuron $i$, thereby verifying the semantic features that the neuron reliably encodes.}
  \label{fig:frame1}
\end{figure*}
\section{Method}
Our goal is to identify and verify neurons in deep neural networks that possess well-defined functions, thereby enabling humans to understand the decision-making mechanisms of these networks better. Previous studies have shown that the semantics encoded by neurons in the penultimate layer~\cite{kalibhat2023identifying} can be interpreted by humans. Therefore, our analysis primarily focuses on the functionality of neurons in these layers. Motivated by scientific methodology, we present a three-stage framework for neuron interpretation and verification (see \cref{fig:frame1} for an overview). The framework consists of three key stages: select samples with consistently high activation (Section \cref{meth:stage1}), generate concept hypotheses from the selected samples (Section \cref{meth:stage2}), and verify the hypothesized concepts (Section \cref{meth:stage3}).

\subsection{Definitions}
\textbf{Definition 1} (Neuron Activation). Given a pretrained classification network $f$ and an input image $x$, the activation value of the $i$-th neuron in layer $l$ is denoted as $a_{i}^{l}(x)$. This value measures the response strength of the neuron when processing the input image.

\noindent\textbf{Definition 2} (Neuron Activation Distribution). For a probe dataset $\mathcal{D}_{probe}$, the activation distribution of neuron $i$ is defined as the set $\left \{ a_{i}^{l}(x) \mid x \in \mathcal{D}_{probe} \right \}$.It describes how the neuron responds to different images in the dataset.
\label{Definition:2}
\subsection{Select High-Activation Samples }
\label{meth:stage1}

\begin{figure}[t]
  \centering
  \includegraphics[width=0.9\linewidth]{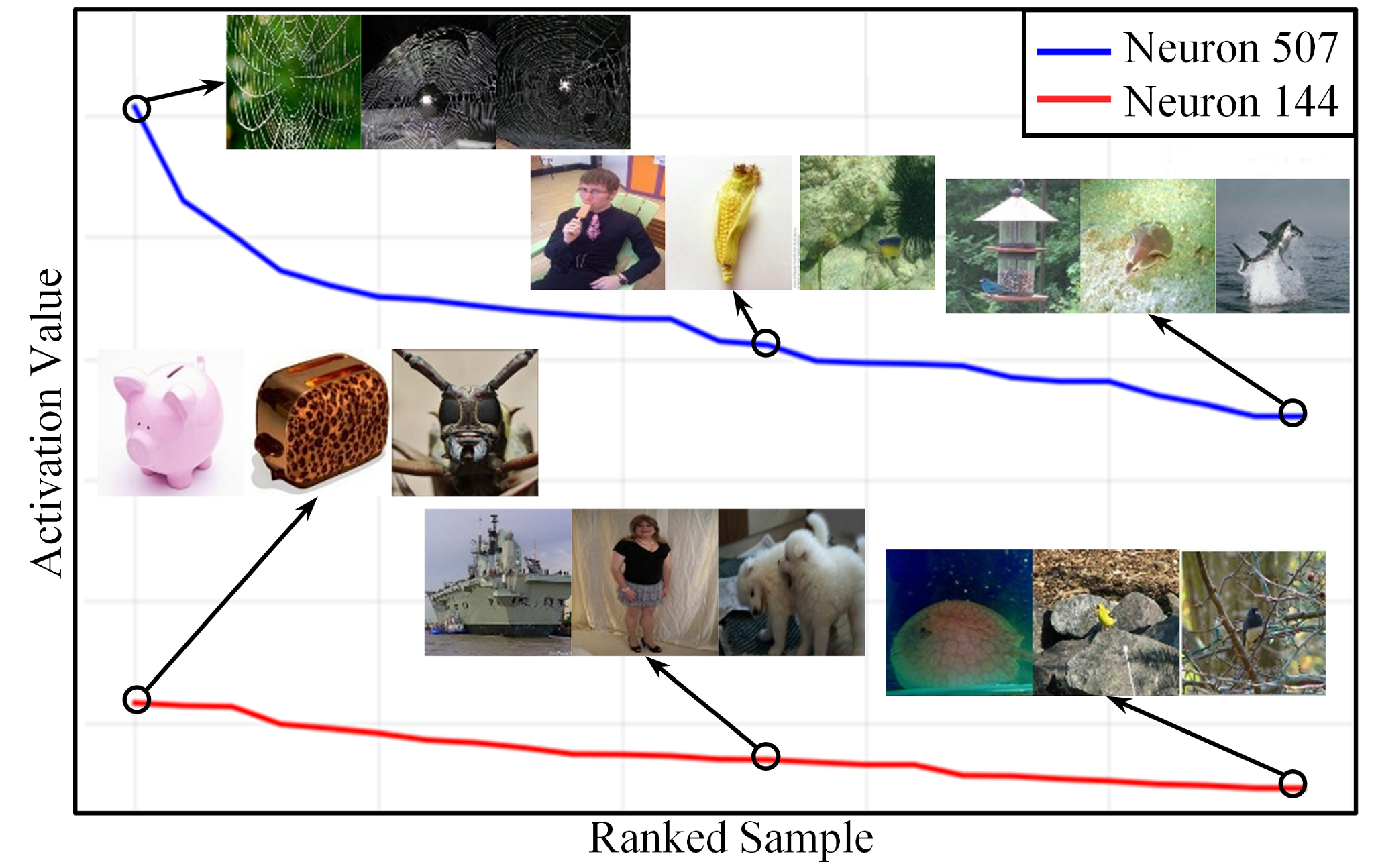}  
  \caption{
    Activation distributions and Concepts of high-discrimination neurons (Neuron 507) and low-discrimination neurons (Neuron 144). The high activation of high-discrimination neurons exhibits consistent patterns while the high activation of low-discrimination neurons fails to do so.
  }
  \label{fig:select}
\end{figure}

We use a sample-selection mechanism to identify neuron samples that are most indicative of well-defined functional roles. As shown in ~\cref{fig:select}, this method distinguishes high-discrimination neurons (e.g., Neuron 507) with consistent activation patterns from low-discrimination neurons (e.g., Neuron 144) with scattered responses.
Specifically, based on \cref{Definition:2}, we can obtain the activation distribution of neuron $i$ over the probe dataset $\mathcal{D}_{probe}$. We then calculate the ratio of the 99-th percentile to the median of this distribution to quantify the neuron’s response. A higher ratio indicates that the neuron exhibits high activation in response to specific stimuli and low activation in response to non-preferred stimuli.
According to the activation distribution of neurons, we define a threshold $\beta$. If a neuron’s ratio exceeds this threshold, it can be considered to encode distinct, well-defined functional features. Through this step, we construct a high-quality set of candidate samples $\mathcal{D}_{i}^{high}=\left \{ x_{1},...,x_{k} \right \}$, which is used in subsequent hypothesis and verification stages. We select the top 20 samples for each neuron, striking a balance between capturing meaningful neural responses and ensuring computational efficiency.



\subsection{Hypothesize Concepts}
\label{meth:stage2}
This stage aims to formulate hypotheses for the candidate sample set $\mathcal{D}_{i}^{high}$, describing the potential functional roles. It consists of two steps: high-activation images are clustered to identify distinct functional patterns, and precise concept hypotheses are independently generated for each cluster.
\begin{figure*}[t]  
  \centering
  \includegraphics[width=0.95\textwidth]{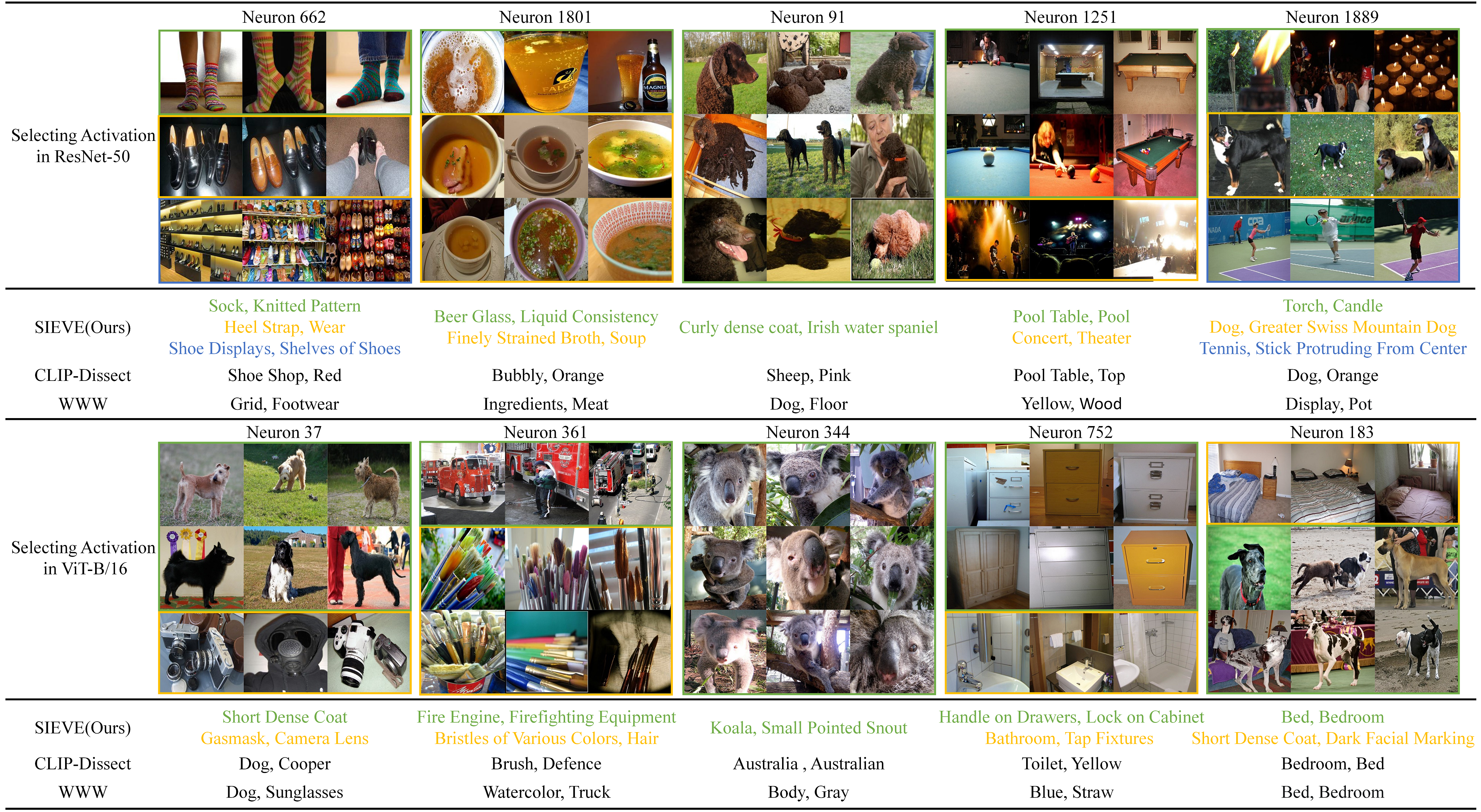}  
  \caption{
    Functional descriptions of penultimate-layer neurons in ResNet-50 and ViT-B/16, generated by SIEVE, CLIP-Dissect~\cite{Dissect} and WWW~\cite{ahn2024unified}. SIEVE provides more complete and accurate semantic explanations. It captures localized features and multiple concepts, rather than only broad attributes such as object category or color. For example, ViT-B/16 neuron 37 is described by SIEVE with specific cues like Short Dense Coat, whereas CLIP-Dissect and WWW may omit some relevant concepts or provide only coarse labels such as Dog. Overall, the neuron-level characterizations produced by SIEVE are more fine-grained and validated.
  }
  \label{fig: qual}
\end{figure*}

\textbf{Cluster.} For neuron $i$, a set of highly activated samples $\mathcal{D}_{i}^{high}$ is obtained as described in \cref{meth:stage1}. 
To focus on the regions genuinely responsive to the neuron, each sample is cropped according to the neuron’s activation map, yielding a corresponding set of image patches $P_{i}^{high}=\left \{ p_{1},...,p_{k} \right \} $, where each $p_{j}$ represents the local region with the highest activation value on the map. 
This procedure effectively removes irrelevant background, ensuring that the subsequent feature distribution better reflects the neuron’s true function. Feature vectors are then extracted from the set of image patches $P_{i}^{high}$, resulting in a feature set $V_{i}^{high}=\left \{ v_{1},...,v_{k} \right \}$. 
The feature set $V_{i}^{high}$ is then clustered to reveal the multiple potential functions a single neuron may correspond to. 
Specifically, agglomerative clustering~\cite{murtagh2014ward} is applied to group the feature vectors extracted from the high-activation patches, with the number of clusters being automatically determined by the Silhouette score~\cite{rousseeuw1987silhouettes}, which evaluates the compactness and separability of the clusters. The clustering results partition the high-activation image patches into $m$ clusters $\left \{ C_{i,1},...,C_{i,m} \right \}$, where $C_{i,j}$ represents the $j$-th typical response pattern of neuron $i$.


\textbf{Hypothesize.}
To describe each cluster $C_{i,j}$, we follow CLIP-Dissect~\cite{Dissect} and introduce a predefined concept set $\mathcal{D}_{concept}$ as $\mathcal{T}=\left \{ t_{1},...,t_{n} \right \} $. We introduce a vision-language model (e.g., CLIP), employing its image encoder $E_{I}(\cdot)$ and text encoder $E_{T}(\cdot)$ to project images and texts into a shared embedding space. For each high-activation image patches $x_{p} \in \mathcal C_{i,j}$ and each concept $t_{q} \in \mathcal{T}$, their similarity is calculated as:
\begin{equation}
  s(x_{p},t_{q})=sim(E_{I}(x_{p}),E_{T}(t_{q}))
  \label{eq:sim}
\end{equation}
where $sim(\cdot, \cdot)$ denotes cosine similarity. The matching score between cluster $C_{i,j}$ and concept $t_{q}$ is represented by the average similarity:
\begin{equation}
  g(t_{q},C_{i,j})=\frac{1}{|C_{i,j}|}\sum_{x_{p} \in C_{i,j}}s\bigl(x_{p},t_{q}\bigr)
  \label{eq:avg sim}
\end{equation}
Concepts with high scores are selected as the functional hypothesis of the cluster (we set $K=2$):

\begin{equation}
h_{i,j} = \operatorname{arg\,top\text{-}}K
\left( \{g(t_q, C_{i,j}) \mid t_q \in \mathcal{T}\}, K \right)
\end{equation}
where $\operatorname{arg\,top\text{-}}K(\cdot, K)$ returns the top $K$ concepts $t_q$ with the highest scores. Thus, neuron $i$ obtains a set of concept labels $\mathcal{H}=\left \{ h_{i,1},...,h_{i,m} \right \} $, where each $h_{i,j}$ represents the functional hypothesis for the neuron’s $j$-th high-activation pattern. This mapping captures the complex functions of neurons more accurately than a single hypothesis.

\subsection{Verify Hypothesized Concepts}
\label{meth:stage3}
To verify the hypothesized concepts, we introduce the Activation Rate (AR) to measure the stability of a neuron’s activation on images generated from the hypothesized concepts. The core idea is as follows: if a hypothesized concept $h_{i,j}$ reflects the functionality of neuron $i$ in cluster $C_{i,j}$, then images generated based on this concept should elicit a strong response from neuron $i$. Unlike traditional destructive interventions, such as neuron ablation, we take a constructive approach to intervention: we actively generate input stimuli that align with the hypothesised concept and observe how the neuron responds, thereby assessing the consistency and reliability of the interpretation. 



\textbf{Hypothesis-Guided Image Generation.}
For each hypothesis concept $h_{i,j}$ to be verified, we use it as a textual prompt for a pretrained text-to-image generation model $G$ (e.g., Stable Diffusion~\cite{DBLP:conf/iclr/PodellELBDMPR24}). By sampling different random noise vectors, we generate a set of images independent of the probing dataset $\mathcal{D}_{probe}$, denoted as $\mathcal {D}_{\text{gen}}^{(i,j)} = \{\, G(h_{i,j}; z) \mid z \sim \mathcal{N}(0, I) \,\}$. These generated samples are primarily determined by the hypothesized concept $h_{i,j}$, which helps minimize potential interference from the probing data or model biases, while may still be influenced by the training data from the generative model.




\begin{table*}[t]
\centering
\caption{Quantitative comparison of neuron interpretation methods of ResNet-50 pre-trained on ImageNet-1K, using the ImageNet-1K validation set as $D_{probe}$. We evaluate the metrics proposed by CLIP-Dissect~\cite{Dissect} in the final layer, and the mean Activation Rate in the penultimate layer. Bold numbers indicate the best results. We use the Common Words as the concept set following CLIP-Dissect.}
\label{tab:Resnet50}
\small
\setlength{\tabcolsep}{12pt} 
\renewcommand{\arraystretch}{0.9}
\begin{tabular}{l|c c|c|c|c} 
\toprule
\textbf{\multirow{2}{*}{Method}} & \multirow{2}{*}{$\bm{\mathcal{D}_{probe}}$} & \multirow{2}{*}{$\bm{\mathcal {D}_{concept}}$}
& \multicolumn{2}{c|}{\textbf{Final Layer}} & \multicolumn{1}{c}{\textbf{Penultimate Layer}} \\
\cmidrule(lr){4-5} \cmidrule(lr){6-6}
 &  &  & \textbf{CLIP cos} & \textbf{mpnet cos} & \textbf{mean AR (\%)} \\ 
\midrule
Network Dissect & Broden & Broden & 0.7073 & 0.3256 & 45.01 \\
MILAN(b) & ImageNet val & -- & 0.7192 & 0.3089 & 46.10 \\
FALCON & ImageNet val & LAION-400m & 0.7031 & 0.2193 & 46.32 \\
DnD & ImageNet val & GPT-3.5 Turbo & 0.7595 & 0.4371 & 51.46 \\
\midrule
\multirow{2}{*}{CLIP-Dissect} & ImageNet val & Broden (1.2k) & \textbf{0.7942} & \textbf{0.4543} & 55.16 \\
                              & ImageNet val & Common Words (3k) & 0.7868 & 0.4462 & 57.91 \\
\midrule
\multirow{2}{*}{WWW} & ImageNet val & Broden (1.2k) & 0.7792 & 0.4327 & 49.55 \\
                     & ImageNet val &  Common Words (3k) & 0.7713 & 0.4463 & 50.23 \\
\midrule
\multirow{2}{*}{Ours} & ImageNet val & Broden (1.2k) & 0.7848 & 0.4452 & \textbf{85.73} \\
                      & ImageNet val &  Common Words (3k) & \textbf{0.7914} & \textbf{0.4547} & \textbf{86.29} \\
\bottomrule
\end{tabular}
\end{table*}

\begin{table*}[t]
\centering
\caption{Quantitative comparison of neuron interpretation methods of ViT-B/16 pre-trained on ImageNet-1K. }
\label{tab:vitb16}
\small
\setlength{\tabcolsep}{13pt} 
\renewcommand{\arraystretch}{0.9}
\begin{tabular}{l|c c|c|c|c} 
\toprule
\textbf{\multirow{2}{*}{Method}} & \multirow{2}{*}{$\bm{\mathcal{D}_{probe}}$} & \multirow{2}{*}{$\bm{\mathcal {D}_{concept}}$}
& \multicolumn{2}{c|}{\textbf{Final Layer}} & \multicolumn{1}{c}{\textbf{Penultimate Layer}} \\
\cmidrule(lr){4-5} \cmidrule(lr){6-6}
 &  &  & \textbf{CLIP cos} & \textbf{mpnet cos} & \textbf{mean AR (\%)} \\ 
\midrule
FALCON & ImageNet val & LAION-400m & 0.7105 & 0.2281 & 45.95 \\
DnD & ImageNet val & GPT-3.5 Turbo & 0.7462 & 0.4354 & 52.47 \\
\cmidrule{1-6}
\multirow{2}{*}{CLIP-Dissect} & ImageNet val & Broden (1.2k) & 0.7651 & \textbf{0.4519} & 56.21 \\
                              & ImageNet val &  Common Words (3k) & 0.7538 & 0.4437 & 57.70 \\
\cmidrule{1-6}
\multirow{2}{*}{WWW} & ImageNet val & Broden (1.2k) & 0.7455 & 0.4261 & 49.19 \\
                     & ImageNet val &  Common Words (3k) & 0.7512 & \textbf{0.4520} & 51.88 \\
\cmidrule{1-6}
\multirow{2}{*}{Ours} & ImageNet val & Broden (1.2k) & \textbf{0.7742} & 0.4401 & \textbf{84.87} \\
                      & ImageNet val &  Common Words (3k) & \textbf{0.7858} & 0.4340 & \textbf{85.24} \\
\bottomrule
\end{tabular}
\end{table*}

\textbf{Activation Consistency Measurement.}
After generating the verification samples, we quantify the consistency of a neuron’s responses to the hypothesised concepts using the activation rate. This allows us to assess the reliability of the functional hypotheses. Specifically, for each neuron $i$ to be verified, its activation distribution over the probing dataset $\mathcal{D}_{probe}$ can be obtained as $\left \{ a_{i}^{l}(x) \mid x \in \mathcal{D}_{probe} \right \}$. Based on this distribution, the activation threshold $T_{i}$ (Top 1\%) is determined to distinguish significant from non-significant activations. For the set of verification samples $\mathcal {D}_{gen}^{(i,j)}$ generated from the hypothesized concept $h_{i,j}$, the Activation Rate of neuron $i$ is defined as:


\begin{equation}
  AR_{i} = \frac{1}{|\mathcal {D}_{gen}^{(i,j)}|} \sum_{x_{gen} \in \mathcal {D}_{gen}^{(i,j)}} 1\{a_i^l(x_{gen})>T_{i}\},
  \label{eq:avg activate}
\end{equation}
where $a_i^l(x_{gen})$ denotes the activation value of neuron $i$ for the generated image $x_{gen}$, and $1\{\cdot\}$ is the indicator function. This metric measures the proportion of generated images, relevant to the neuron’s functional hypothesis, that elicit significant activation, reflecting the stability and consistency of the hypothesized function. After computing the Activation Rate for each neuron, we calculate the overall mean to obtain the global consistency, referred to as the mean Activation Rate. The initial mean Activation Rate is used as a filtering criterion, where hypotheses below this value are discarded, and the mean is recomputed over the retained set. This metric quantifies the alignment between the neuron's functional interpretations provided by the explanation method and the neuron's true functionality, and can be used to evaluate the quality of the explanation method.

\section{Experiment}
We conduct extensive experiments to evaluate the proposed framework. \cref{experiment: qualitative} and \cref{experiment: Quantitative} present the results from qualitative and quantitative perspectives, respectively; \cref{experiment: Ab} assesses the contribution of each component of the framework through ablation studies; ~\cref{experiment: ATA} investigates the effect of the threshold $\beta$ on the framework; and ~\cref{experiment: Dis} discusses the impact of domain shift on mean AR. Our evaluation covers a variety of models, including ResNet-18~\cite{he2016identity}, ResNet-50~\cite{he2016identity}, and ViT-B/16~\cite{dosovitskiy2020image}. To ensure both coverage and comparability, we adopt the concept set $\mathcal{D}_{concept}$ from CLIP-Dissect~\cite{Dissect} and DEAL~\cite{li2024deal}. The analysis focuses on penultimate layer features of the networks, as they capture high-level semantic representations that are more likely to align with human-interpretable concepts~\cite{kalibhat2023identifying}.




\begin{table*}[t]
\centering
\caption{Quantitative comparison of neuron interpretation methods of ResNet-18 pre-trained on Places365, using the Places365 test set as $D_{probe}$. We evaluate the metrics proposed by CLIP-Dissect~\cite{Dissect} in the final layer, and the mean Activation Rate in the penultimate layer. Bold numbers indicate the best scores under the same settings.}
\label{tab:place365}
\small
\setlength{\tabcolsep}{12pt} 
\renewcommand{\arraystretch}{0.9}
\begin{tabular}{l|c c|c|c|c} 
\toprule
\textbf{\multirow{2}{*}{Method}} & \multirow{2}{*}{$\bm{\mathcal{D}_{probe}}$} & \multirow{2}{*}{$\bm{\mathcal {D}_{concept}}$}
& \multicolumn{2}{c|}{\textbf{Final Layer}} & \multicolumn{1}{c}{\textbf{Penultimate Layer}} \\
\cmidrule(lr){4-5} \cmidrule(lr){6-6}
 &  &  & \textbf{CLIP cos} & \textbf{mpnet cos} & \textbf{mean AR (\%)} \\ 
\midrule
FALCON & ImageNet val & LAION-400m & 0.6975 & 0.2203 & 47.15 \\
DnD & Places365 test & GPT-3.5 Turbo & 0.7458 & 0.4419 & 53.12 \\
\cmidrule{1-6}
\multirow{2}{*}{CLIP-Dissect} & Places365 test & Broden (1.2k) & 0.7761 & 0.4462 & 58.32 \\
                              & Places365 test &  Common Words (3k) & \textbf{0.7883} & \textbf{0.4528} & 56.90 \\
\cmidrule{1-6}
\multirow{2}{*}{WWW} & Places365 test & Broden (1.2k) & 0.7617 & \textbf{0.4587} & 52.23 \\
                     & Places365 test &  Common Words (3k) & 0.7526 & 0.4287 & 46.79 \\
\cmidrule{1-6}
\multirow{2}{*}{Ours} & Places365 test & Broden (1.2k) & \textbf{0.7936} & 0.4453 & \textbf{85.82} \\
                      & Places365 test & Common Words (3k) & 0.7833 & 0.4355 & \textbf{84.40} \\
\bottomrule
\end{tabular}
\end{table*}

\subsection{Qualitative Results}
\label{experiment: qualitative}
We qualitatively evaluated the effectiveness of SIEVE in providing neuron-level concept interpretations, focusing on the penultimate layer neurons of two representative architectures: ResNet-50 and ViT-B/16. For each neuron, we present its most highly activated images alongside textual concept explanations generated by SIEVE, CLIP-Dissect, and WWW, as illustrated in ~\cref{fig: qual}. In general, SIEVE demonstrates greater alignment and granularity in concept interpretation, producing descriptions that are closely aligned with neuron activations and capable of capturing local features and diverse patterns. In contrast, baselines often yield coarse or incorrect labels. SIEVE generates rich and precise concepts across different architectures. For instance, for Neuron 361 in ViT-B/16, SIEVE offers two distinct descriptions, whereas baseline methods capture only one. In particular, SIEVE focuses on common features across highly activated images, such as “Bristles of Various Colors”, while baseline methods tend to assign more specific object categories, such as “Brush”. This suggests that SIEVE can provide detailed and trustworthy insights into neuron behavior.

\subsection{Quantitative Results}
\label{experiment: Quantitative}
We evaluate the performance of the proposed method compared with baselines through quantitative experiments. Following the metrics introduced by CLIP-Dissect~\cite{Dissect}, we measure the effectiveness of a method by computing the similarity between each neuron's hypothesized concept in the final layer and ground-truth labels. However, these metrics only capture the consistency between the functional descriptions of neurons in the final layer and ground-truth labels. They cannot assess the accuracy of neuron function descriptions in the penultimate layer. To this end, we use the mean Activation Rate, which does not rely on ground-truth labels and quantifies the consistency between a neuron’s functional description produced by an interpretability method and the actual function of the neuron, enabling the evaluation of neuron functions beyond the final layer.

We conducted quantitative experiments on ImageNet-pretrained ResNet-50 and ViT-B/16 models, comparing our approach with Network Dissection, CLIP-Dissect~\cite{Dissect}, MILAN(b)~\cite{hernandez2021natural}, FALCON~\cite{kalibhat2023identifying}, WWW~\cite{ahn2024unified} and DnD~\cite{bai2025interpreting}. As shown in \cref{tab:Resnet50} and \cref{tab:vitb16}, we report two metrics proposed by CLIP-Dissect as well as the mean Activation Rate. Results show that our method achieves the best or second-best performance in most cases. Specifically, on both ResNet-50 and ViT-B/16, our approach outperforms baselines in CLIP and MPNet similarity, indicating more accurate concept descriptions. More importantly, our method shows a clear advantage on the mean Activation Rate, further validating that the “Select–Hypothesize–Verify” framework generates neuron concepts with high semantic relevance. To evaluate the generalizability of our method, we tested it on ResNet-18 pretrained in Places365 (\cref{tab:place365}), showing that our approach consistently achieves top performance even on a different dataset.

\subsection{Ablation Study}
\label{experiment: Ab}
To verify the necessity of each component in our framework, we conduct an ablation study using the ResNet-50 pretrained on ImageNet. The baseline does not apply the activation threshold $\beta$, clustering, or concept verification. As shown in \cref{tab:Ablation}, the Verify module has the greatest impact, indicating that validating neuron–concept hypotheses is crucial for ensuring a reliable semantic interpretation. Select improves the consistency of interpretations by selecting samples that contribute stable and discriminative neuron activations, effectively filtering out less informative or ambiguous cases. Clustering may serve a similar purpose to selection. Once all the components have been brought together, the framework achieves the greatest possible alignment across all the metrics, thereby demonstrating the complementary effects of selection, clustering, and verification.

\begin{table}[t]
\centering
\caption{Ablation study on Select, Cluster, and Verify components. Results show that all modules contribute positively to interpretation quality, with the full framework achieving the best performance across CLIP cos, mpnet cos and mean Activation Rate.}
\label{tab:Ablation}
\small
\setlength{\tabcolsep}{1.4pt} 
\renewcommand{\arraystretch}{0.9}
\begin{tabular}{c c c|c|c|c} 
\toprule
\multirow{2}{*}{\textbf{Select}} & \multirow{2}{*}{\textbf{Cluster}} & \multirow{2}{*}{\textbf{Verify}} 
& \multicolumn{2}{c|}{\textbf{Final Layer}} & \multicolumn{1}{c}{\textbf{Penultimate Layer}} \\
\cmidrule(lr){4-5} \cmidrule(lr){6-6}
 &  &  & \textbf{CLIP cos} & \textbf{mpnet cos} & \textbf{mean AR (\%)} \\
\midrule
 &  &  & 0.6738 & 0.2306 & 45.57 \\
 & \checkmark & \checkmark & 0.7624 & 0.4301 & 77.90 \\
 \checkmark &  & \checkmark & 0.7821 & 0.4423 & 81.52 \\
 \checkmark & \checkmark &  & 0.7656 & 0.4189 & 72.87 \\
\checkmark & \checkmark & \checkmark & \textbf{0.7914} & \textbf{0.4547} & \textbf{86.29} \\
\bottomrule
\end{tabular}
\end{table}

\subsection{Activation Threshold Analysis}
\label{experiment: ATA}
In our framework, when selecting high-activation samples, we take into account not only the magnitude of individual activations but also their ratio relative to the median of the neuron’s overall activation distribution. Some samples, even though they rank among the top activations, have a low ratio relative to this median, which suggests that the neuron lacks a clear activation preference. The hyperparameter $\beta$ helps reduce the number of samples to be processed, thereby focusing on those that provide discriminative information for interpretation. Experiments were conducted using the ResNet-50 model pretrained on ImageNet. As shown in \cref{tab:β}, varying $\beta$ has minimal impact on the final interpretability metrics, since the most semantically informative samples are consistently retained. The $\beta$ can be used to control sample quality without compromising the reliability of neuron-level semantic interpretations.

\begin{table}[h]
\centering
\caption{Performance under different activation thresholds $\beta$. The model achieves the best results at 10, which is therefore adopted throughout the experiments in this paper.}
\label{tab:β}
\small
\setlength{\tabcolsep}{9.5pt} 
\renewcommand{\arraystretch}{0.9}
\begin{tabular}{c|c|c|c} 
\toprule
\multirow{2}{*}{\textbf{$\beta$}} & \multicolumn{2}{c|}{\textbf{Final Layer}} & \textbf{Penultimate Layer} \\
\cmidrule(lr){2-3} \cmidrule(lr){4-4}
 & \textbf{CLIP cos} & \textbf{mpnet cos} & \textbf{mean AR (\%)} \\
\midrule
4 & 0.7932 & 0.4421 & 86.05 \\
6 & \textbf{0.7987} & 0.4517 & 85.61 \\
8 & 0.7805 & 0.4432 & 85.72 \\
10 & 0.7914 & \textbf{0.4547} & \textbf{86.29} \\
12 & 0.7821 & 0.4478 & 85.98 \\
\bottomrule
\end{tabular}
\end{table}


\subsection{Discussion}
\label{experiment: Dis}

\begin{figure}[t]
  \centering
  \includegraphics[width=1.0\linewidth]{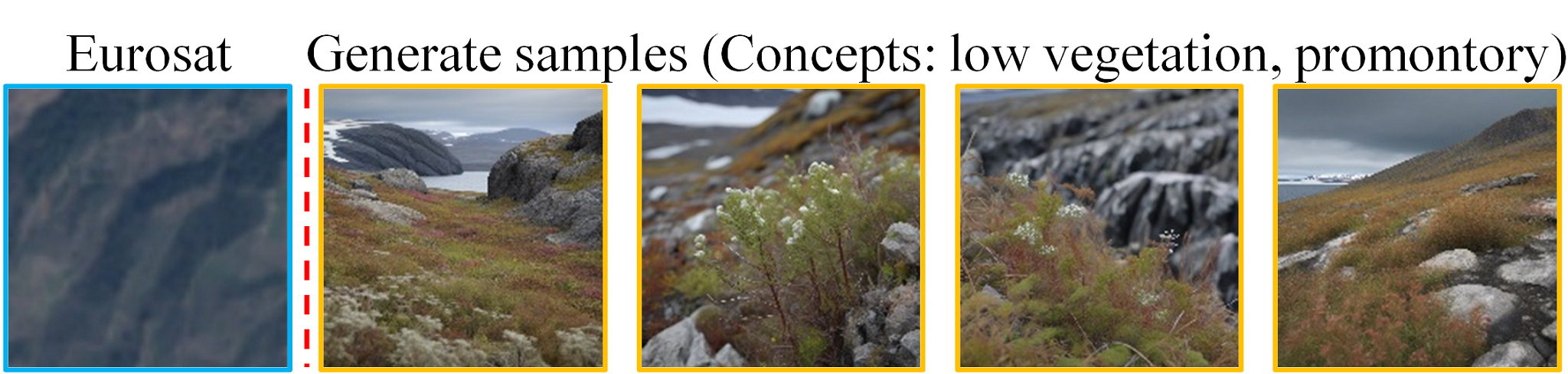}  
  \caption{
    Visualization of the domain shift effect. Significant discrepancies exist between aerial and generated samples.
  }
  \label{fig:domain shift}
\end{figure}

\begin{figure}[t]
  \centering
  \includegraphics[width=1.0\linewidth]{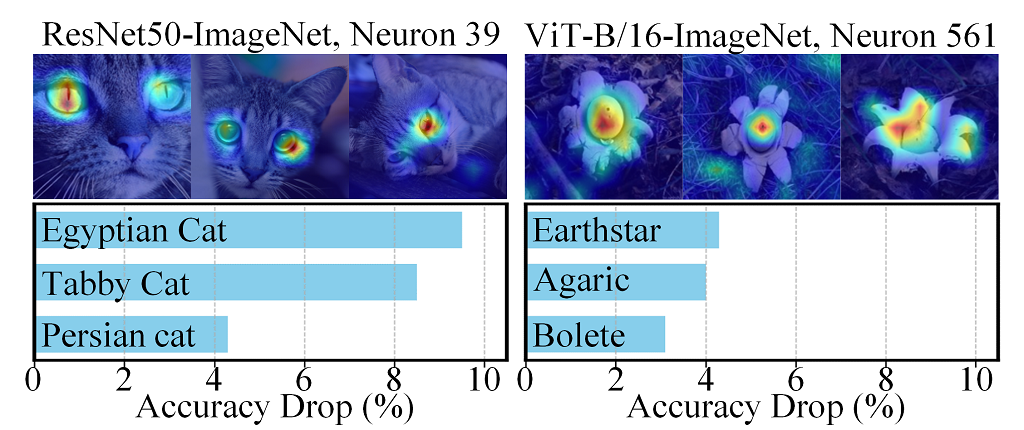}  
  \caption{
    The heatmap (top) shows the part of the image that activates Neuron 39 the most, which is the cat's eyes. The bar plot (bottom) quantifies the ablation effect, showing a targeted accuracy drop for semantically related feline categories.
  }
  \label{fig:keyNeuron}
\end{figure}

To quantify the effect of the domain shift, the ResNet-18 trained on remote sensing data is introduced. As shown in \cref{tab:domain shift}, all methods exhibit degraded results due to the impact of domain shift on the hypothesis and verification. But the verification can keep positive gains under domain shift, mitigating the impact on the hypothesis. The original and generated images are shown in \cref{fig:domain shift}. The significant visual discrepancy in perspective and texture between the aerial view (left) and the generated images (right) intuitively illustrates the challenge introduced by the domain shift. Beyond domain shift, we explore how individual neuron aligns with specific concepts. We ablate individual neurons to measure class-specific accuracy drops. As shown in \cref{fig:keyNeuron}, Neuron 39 in ResNet-50 primarily responds to cat eyes, and ablating it predominantly degrades accuracy for feline categories. A similar pattern is observed in ViT-B/16.


\begin{table}[t]
\centering
\caption{Performance under domain shift. Eurosat represents a distribution significantly different from the generative model's training data, while Places365 represents a general domain.}
\label{tab:domain shift}
\small
\setlength{\tabcolsep}{2pt} 
\renewcommand{\arraystretch}{0.9}
\begin{tabular}{c|c c|c c} 
\toprule
\multirow{2}{*}{\textbf{Method}} & \multicolumn{2}{c|}{\textbf{Eurosat (domain shift)}} & \multicolumn{2}{c}{\textbf{Places365 (general)}} \\
\cmidrule(lr){2-3} \cmidrule(lr){4-5}
 & \textbf{CLIP cos}  & \textbf{mean AR (\%)} & \textbf{CLIP cos}  & \textbf{mean AR (\%)}\\
\midrule
CLIP-Dis. & 0.7152      & 43.16   &   0.7883    &     56.90 \\  
WWW      & 0.7047       & 37.43   &   0.7526    &     46.79 \\  
Ours     & 0.7393       & 75.45   &   0.7833    &     84.40 \\  
\bottomrule
\end{tabular}
\end{table}

\section{Conclusion}
This work focuses on addressing the limitations of current methods, which are limited by the two assumptions: 1) assuming that each neuron can capture discriminative features; and 2) assuming that each concept generated by models is accurate. To address both issues, we propose a Select–Hypothesize–Verify framework for interpreting neuron functionality (dubbed SIEVE), which both filters out low-discriminatory neurons and enables closed-loop verification on neuron concepts to avoid mismatched concepts.




\section*{Acknowledgements}
This work was supported in part by the National Science Foundation of China Joint Key Project under Grant U24B20173, U24B20182, and by the National Natural Science Foundation of China under Grant 62576060, 62536002, 62561160098.

{
    \small
    \bibliographystyle{ieeenat_fullname}
    \bibliography{main}
}


\end{document}